\documentclass[10pt,twocolumn,letterpaper]{article}
\usepackage{hyperref}
\hypersetup{
 linktocpage=true,
 pdfborderstyle={/S/S/W 1},
 hyperindex=true,
 bookmarks=false,
 colorlinks=true,
 breaklinks=true,
 letterpaper=true,
}
\usepackage{subcaption}  
\usepackage{iccv}
\usepackage{times}
\usepackage{epsfig}
\usepackage{graphicx}
\usepackage{amsmath}
\usepackage{amssymb}
\newcommand{\R}{\mathbb{R}}
\usepackage{mathtools}
\usepackage[numbers,sort]{natbib} 
\usepackage{url}

\usepackage{amsmath,amsfonts,bm}

\def\eqref#1{equation~\ref{#1}}

\def\1{\bm{1}}

\def\rvx{{\mathbf{x}}}

\DeclareMathAlphabet{\mathsfit}{\encodingdefault}{\sfdefault}{m}{sl}
\SetMathAlphabet{\mathsfit}{bold}{\encodingdefault}{\sfdefault}{bx}{n}

\usepackage{float, placeins}
\usepackage[symbol]{footmisc}

\iccvfinalcopy 

\def\iccvPaperID{1} 

\ificcvfinal\fi

\begin{document}
\title{Count, Crop and Recognise: Fine-Grained Recognition in the Wild}
\author{Max Bain\textsuperscript{1$\dagger$}, Arsha Nagrani\textsuperscript{1}, Daniel Schofield\textsuperscript{2} and Andrew Zisserman\textsuperscript{1}\\
\\
\textsuperscript{1}Visual Geometry Group, Department of Engineering Science, University of Oxford\\
\textsuperscript{2}Institute of Cognitive \& Evolutionary Anthropology, University of Oxford\\
}

\maketitle
\ificcvfinal\thispagestyle{empty}\fi

\begin{abstract}

The goal of this paper is to label all the animal individuals present in every frame of a video. 
Unlike previous methods that have principally concentrated on labelling face tracks, we aim to label individuals even when their faces are not visible. We make the following contributions: (i) we introduce a `Count, Crop and Recognise' (CCR) multi-stage recognition process for frame level
labelling. The Count and Recognise stages involve specialised CNNs for the task, and we show that this simple staging gives a substantial
boost in performance; (ii) we compare the recall using frame based labelling to both face and body track based labelling,
and demonstrate the advantage of frame based with CCR for the specified goal; (iii) we introduce a new
dataset for chimpanzee recognition in the wild; and (iv) we apply a high-granularity visualisation technique to further understand the learned CNN features for the recognition of chimpanzee individuals.


\end{abstract}
\vspace{-0.5em}
\section{Introduction}
\footnotetext{\textsuperscript{$\dagger$}Correspondence at maxbain@robots.ox.ac.uk}
\label{sec:intro}


Recognising animal individuals in video is a key step towards monitoring the movement, population, and complex social behaviours of endangered species. Traditional individual recognition pipelines rely extremely heavily on the detection and tracking of the face or body, both for 
humans~\cite{Everingham06a,Cour09,Bojanowski13,Kostinger11,Wohlhart11,Yang05,haurilet16,Tapaswi12,Nagrani17b} and for other
species~\cite{deb2018face,witham2018automated,rakotonirina2018role,chimp_bias}. This can be a daunting annotation task, especially for large video corpora of non-human species where custom detectors must be trained and expert knowledge is required to label individuals.
Furthermore, often these detectors fail for animal footage in the wild due to the occlusion of individuals, varying lighting conditions and highly deformable bodies.

\begin{figure*}
    \centering
    \includegraphics[width=1\textwidth]{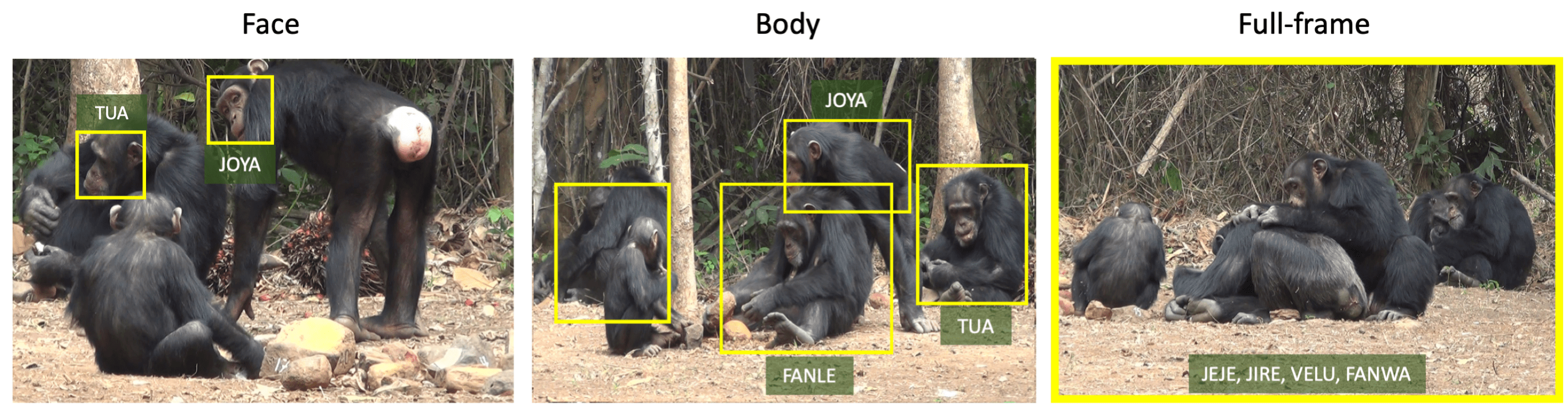}
\caption{\small{Levels of localisation that can be used to recognise individuals in raw footage. Left to right: (1) ~\textbf{Face}: high precision, but often individuals are not detected. (2)~\textbf{Full Body}:
while recall is higher, bodies can be incredibly
difficult to detect due to their extremely deformable
nature. (3)~\textbf{Full Frame}: In this work we explore an
architecture to recognise individuals using only frame level
supervision.}}  \label{fig:teaser}
\vspace{-0.9em}
\end{figure*}

Our goal in this paper is to automatically label individuals in every frame of a video;  but
to go beyond face and body recognition, and  explore identification using the entire frame. 
In doing so we analyse the important
trade off between precision and recall for face, body and full-frame methods for recognition of individuals in video. 
We target the recognition of chimpanzes in the wild.
Consider the performance of models at the three levels of face, body and frame (Figure~\ref{fig:teaser}).
Face recognition now achieves very high accuracy~\cite{schroff2015facenet,Parkhi15,sun2015deepid3} 
for humans due to the availability of
very large datasets for training face detection~\cite{yang2016wider,yan2014face,jain2010fddb,ramanan2012face} and recognition~\cite{kemelmacher2016megaface,guo2016ms,whitelam2017iarpa,bansal2016umdfaces,Cao18}. The result is that the {\em precision} of recognising individuals
will be high, but the {\em recall} may well be low, since, as mentioned above, face recognition will fail for many frames
where the face is not visible. Using a body level model occupies a middle ground between face and frame level:
it offers the possibility of recognising the individual when the head is occluded, e.g.\ by distinguishing marks or shapes
in the case of animals, or by hair or clothes in the case of humans (albeit it is worth noting that changes in clothing
can reduce this advantage --  animals obligingly are unclothed). However, body detectors do not as yet have the same
performance as face detectors, as animal bodies in particularly are highly deformable and can often overlap each other. This means
that bodies may be missed in frames, especially if they are small.  A frame level model offers the possibility of very
high recall (since there are no explicit detectors that can fail, as there are for faces and bodies). In addition, such a method can implicitly use higher-level features for recognition, such as the co-occurrence and spatial relationships between animal individuals (eg. infants are often present in close proximity to the mother). However, the precision
may be low because of the challenge of the large proportion of irrelevant
information present in the frame (in the case of body and particularly face detection, irrelevant information is removed). 

\begin{figure*}
    \centering
    \includegraphics[width=0.8\linewidth]{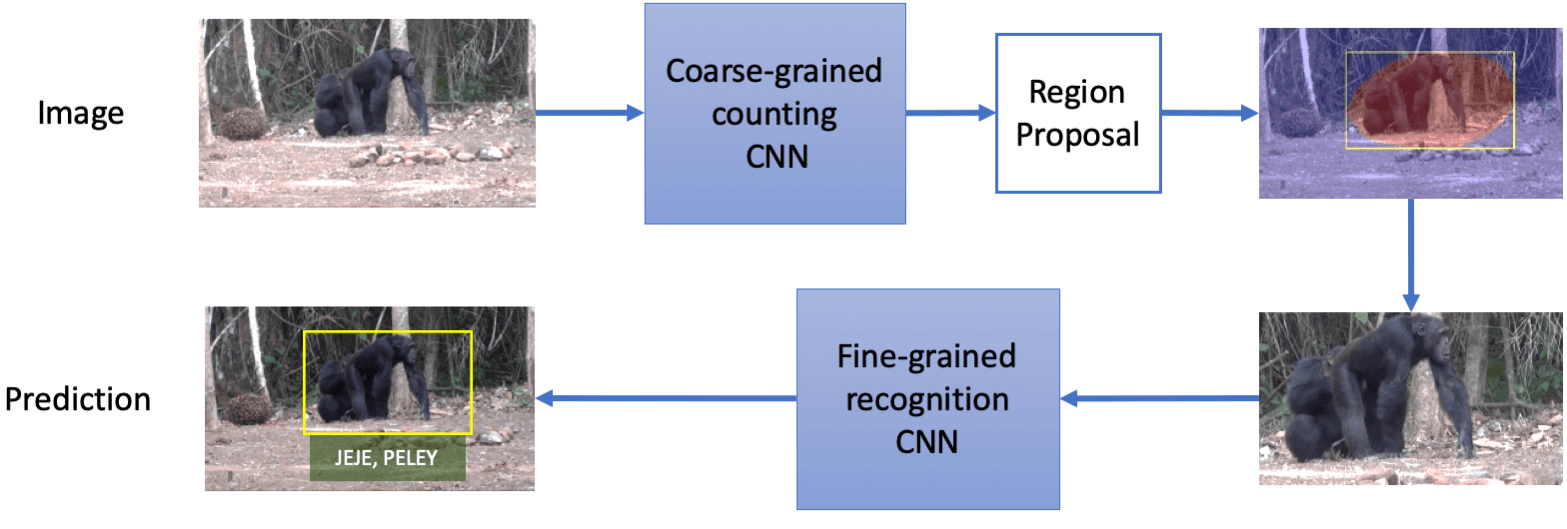}
    \caption{The ~\textit{Count, Crop and Recognise} pipeline consists of three stages: (1) a coarse-grained counting network to count the number of individuals per frame, (2) a crop stage where the class activation maps from the counting network are used to localise regions of interest in the image, and (3) a fine-grained classifier trained on these cropped images.}
    \label{fig:arch}
    \vspace{-0.9em}
\end{figure*}

In this paper we show that the performance of frame level models can be considerably improved by automatically
zooming in on the regions containing the individuals. 
This then enables the best of both worlds: cheap supervision at the frame level, obviating the necessity to train
and employ face or body detectors, and high recall; but with the precision comparable to face and body detection.
We make the following contributions:
(i) we propose a
multi-stage~\textit{Count, Crop and Recognise} (CCR) pipeline to recognise individuals
from raw video with only frame level identity labels required for training. The first two {\em Count and Crop} stages proposes a rectangular region that
tightly encloses all the individuals in the frame. 
The final {\em Recognise} stage
then identifies the individuals in the frame using a multilabel classifier on the rectangular region {\em at full resolution} (Figure~\ref{fig:arch}).
(ii) 
We analyse the trade-offs between using our frame level model and other varying levels of localised supervision for fine-grained individual recognition (at a face, body and frame level) and their respective performances. 
(iii) We have annotated a large, `in the wild' video dataset
of chimpanzees with labels for multiple levels of supervision (face tracks, body tracks, frames) which is available at \texttt{TBD}. 
Finally, (iv) we apply a high-granularity visualisation technique to further understand the learned CNN features for the recognition of chimpanzee individuals.

\section{Related Work}
\noindent\textbf{Animal recognition in the wild:}
Video data has become indispensable in the study of wild animal species ~\cite{caravaggi2017review,nishida2010chimpanzee}. However, animals are difficult objects to recognise, mainly due to their deformable bodies and frequent self occlusion ~\cite{afkham2008joint,berg2006animals}.
Further, variations in lighting, other individual flora, and motion blur create additional challenges. 
Taking inspiration from computer-vision based systems for humans, previous methods for species identification have focused on faces, for chimpanzees~\cite{deb2018face,freytag2016chimpanzee}, tigers~\cite{kuhl2013animal,amur_tigers},
lemurs~\cite{crouse2017lemurfaceid} and even pigs~\cite{hansen2018towards}. Compared to bodies, faces are less deformable and have a fairly standard structure. However, unlike human faces or standard non-deformable object categories, there is a dearth of readily available detectors that can be used off the shelf to localize animals in a frame, requiring researchers to annotate datasets and train their own detectors. It is also often not clear which part of the animal is the most discriminative, e.g. for elephants ears are commonly used~\cite{douglas1972ecology}, whereas for other mammals unique coat patterns such as stripes for zebras and tigers~\cite{amur_tigers} and spots on Jaguars could be key for recognition~\cite{harmsen2017long}. Moving to a full-frame method obviates the need to identify a key discriminating region.
Popular wildlife recognition datasets, such as iNaturalist~\cite{van2018inaturalist}, contain species level labels and
in contrast to our dataset, typically contain a \textit{single} instance of a class clearly visible in the foreground. While a valuable dataset does exist for the individual recognition of chimpanzees~\cite{freytag2016chimpanzee,loos2012detection}, this dataset only contains cropped faces of individuals from zoo enclosures, less applicable to applications of conservation in the wild. \\

\noindent\textbf{Human recognition in TV and film videos:}
The original paper in this area by Everingham {\it et
al.}~\cite{Everingham06a} introduced three ideas: (i) associating 
faces in a shot using tracking by detection, so that a face-track is
the `unit' to be labelled; (ii) the use of aligned transcripts with
subtitles to provide supervisory information for character labels;
and (iii) visual speaker detection to strengthen the supervision (if a
person is speaking then their identity is known from the aligned
transcript). 
Many others have adopted and extended these ideas.
Cour {\it et al.}~\cite{Cour09} cast the problem as
one of ambiguous labelling. Subsequently, 
Multiple
Instance Learning (MIL), was employed by~\cite{Bojanowski13,Kostinger11,Wohlhart11,Yang05,haurilet16}.  Further 
improvements include:
unsupervised and partially-supervised metric
learning~\cite{Cinbis11,Guillaumin10}; 
the range of face viewpoints used 
(e.g.\ adding profile face tracks in addition to the original near-frontal face tracks)~\cite{Everingham09,Sivic09}; and obtaining an episode
wide consistent labelling~\cite{Tapaswi12} (by using a graph
formulation and other visual cues). 
Recent work~\cite{Nagrani17b} has explored using {\em only} face and voice recognition, without the use
of weak supervision from subtitles. \\
\\
\noindent\textbf{Frame level supervision:}
The task of labelling image regions given only frame level labels is that of weakly supervised segmentation: every image is known to have (or not) -- through the image (class) labels -- one or several pixels matching the label.  However, the positions of these
pixels are unknown, and have to be inferred. Early deep learning works on this area include~\cite{Pinheiro15,Pathak_2015_ICCV,Kolesnikov16}. Our problem
differs in that it is fine-grained -- all the object classes are
chimpanzees that must be distinguished, say, rather than the 20 
PASCAL VOC classes of~\cite{Pinheiro15,Pathak_2015_ICCV,Kolesnikov16}. While there have been works on localising fine-grained objects with weak supervision \cite{bilen2016weakly,gao2018c,hu2019see}, they deal only with the restricted case of one instance per image (i.e.\ an image containing a single bird of class \textit{Horned Puffin}). As far as we know, we are the first to tackle the challenging task of classifying multiple fine-grained instances in a single frame with weak supervision. 


\FloatBarrier
  
\section{Count, Crop and Recognise (CCR)}
\label{sec:ccr}
\begin{figure}[h]
\vspace{-0.9em}
    \centering
    \includegraphics[width=1\columnwidth]{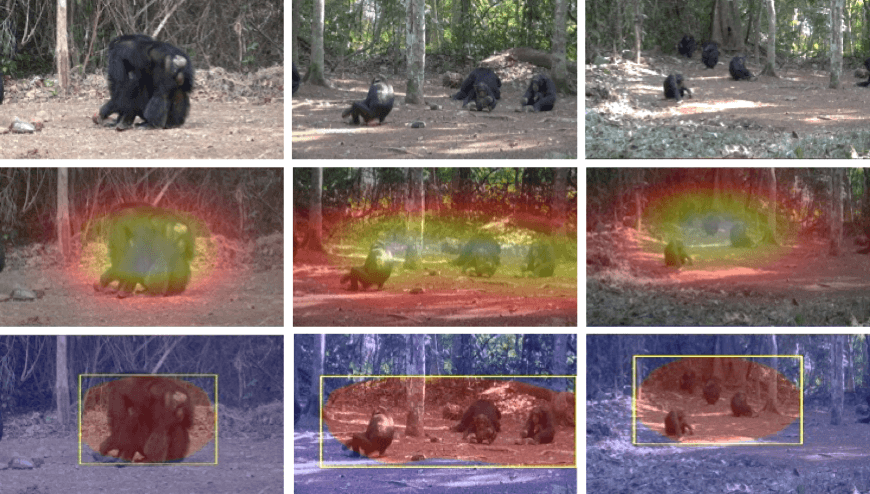}
    \caption{Region proposals for the Chimpanzee Bossou dataset. These are learnt via our counting CNN with no detection supervision at all. Top row: original frame; middle row: CAM for the count; bottom row: region proposal. Note in the second column, how the localisation works well even when the individuals are far apart from each other.}
    \label{fig:proposals}
    \vspace{-0.7em}
\end{figure}

Our goal is, given a frame of a video, to predict all the individuals present in that frame. We would like to learn to do this task with only~\textit{frame-level} labels, i.e no detections and hence no correspondences (who's who). 
The major challenge with such a method, however is that frames contain a lot of irrelevant background noise (Figure~\ref{fig:proposals}), and the distinctions between different individuals is often very fine-grained and subtle (these fine details are hard to learn due to the limited input resolution of CNNs). 

Hence we propose a multi-stage, frame level pipeline that automatically crops discriminative regions containing individuals and so eliminates as much background information as possible, while maintaining the high resolution of the original image. This is achieved by training a deep CNN with a coarse-grained counting objective (a much easier task than fine-grained recognition), before performing identity recognition. The method is loosely inspired by the weakly-supervised object detection method C-WSL~\cite{gao2018c}, 
however, unlike this work, our method requires neither explicit object proposals nor an existing weakly supervised detection method.
Since we do not require exact bounding boxes~\textit{per instance}, but simply a generic zoomed in region, we use class guided activation maps to determine the region of focus. The multiple stages of our CCR method are described in more detail below. Precise implementation details can be found in Section \ref{sec:implement}, and a diagrammatic representation of the pipeline can be seen in Figure. \ref{fig:arch}.


Let $\rvx \in \R^{C \times H \times W}$ be a single frame of the video and let $Y \in \{0,1\}^k$ be a finite vector denoting which of the total $k$ individuals are visible. $Y[i] = 1$ if the $i$-th individual is visible in $\rvx$, and $Y[i] = 0$ otherwise.

\noindent\textbf{Count:} We first train a parameterised function $ c_{\theta}(\rvx')$, given a resized image input $\rvx' \in \R^{C \times H' \times W'}$ to count the number of individuals $n$ within a frame.
In general, we can cast this problem as either a multiclass problem or a regression problem. Since the number of individuals per frame in our datasets is small, we pose this counting task as one of multiclass classification, where the total number of individuals present can be categorised into one of the following classes $ n \in \{0, 1, ..., N\} $ where all counts of $N$ or more are binned into a single bin, with $N$ selected as a hyperparameter (in this work we use $N=3$). The `Negatives' class ($n=0$) is very important for training. Labels for counting come for free with frame level annotation (total number of labels per frame, or $n = |Y|$).  The loss to be minimised can then be framed as a cross-entropy loss on the target $n$ values. In this work we instantiate $c(\rvx')$ as a deep convolutional neural network (CNN) with convolutional layers followed by fully connected layers. Generally $H', W' < H, W$ due to the discrepancy in resolution of raw images and pretrained CNNs.\\
\\
\noindent\textbf{Crop:} Class Activation Maps (CAMs)~\cite{zhou2016learning} are generated from the counting model $c_{\theta}(\rvx')$ to localise the discriminative regions. For resized input image $\rvx'$, let $f_{k}(i,j)$ denote the activation of a unit $k$ in the last convolutional layer, and $w_{k}^{n}$ denote the weight corresponding to count $n$ for unit $k$. The CAM, $M_n$, at each spatial location is given by:
\vspace{-0.8em}
\begin{equation}
M_{n}(i,j) = \sum_{k}{w_{k}^{n}f_{k}(i,j)}
\end{equation}

describing the importance of visual patterns at different spatial locations for a given class, in this case a count. By upsampling the CAM to the same size of $\rvx$ ($H, W$) image regions most relevant to the particular category can be identified. The CAM is then normalised and segmented:
\vspace{-0.5em}
\begin{equation}
M_{n}^{norm}(i,j) = \frac{M_{n}(i,j) - \displaystyle\min_{i,j}{M_{n}(i,j)}}{\displaystyle\max_{i,j}{M_{n}(i,j)}}
\end{equation}
\vspace{-0.2em}
\begin{equation}
M_{n}^{thresh}(i,j) =
\begin{dcases}
    1,& \text{if } M_{n}^{norm}(i,j) > T\\
    0,              & \text{otherwise}
\end{dcases}
\end{equation}

where $T \in [0,1]$ is the chosen threshold value. The largest connected component in $M_{n}^{thresh}$ is found using classical component labelling algorithms \cite{optim_component_labelling, component_labelling}, examples shown in Figure~\ref{fig:proposals}. The bounding box enclosing this component is used to crop the original input image $\rvx$ to get $\rvx_{crop}$, removing irrelevant portions of the image and permitting higher resolution of the cropping region. \\

\noindent\textbf{Recognise:} The cropped regions $\rvx_{crop}$ are used to train a fine grained recognition classifier $R_{\phi}(\rvx_{crop}')$ using the original frame-level labels $Y$. This second recognition classifier is also instantiated as a CNN, with different parameters $\phi$, and trained for the task of multilabel classification, with one class for every individual in the dataset. We use a weighted Binary Cross-Entropy loss, where the weight $w_i$ for each class $i$ is: $w_i = f_{max} / f_{i}$, where $f_{max}$ refers to the number of instances for the most populous class, and $f_i$ is the number of instances for class $i$. \\

\begin{figure}
  \vspace{-0.5em}
    \centering
    \includegraphics[width=1\columnwidth]{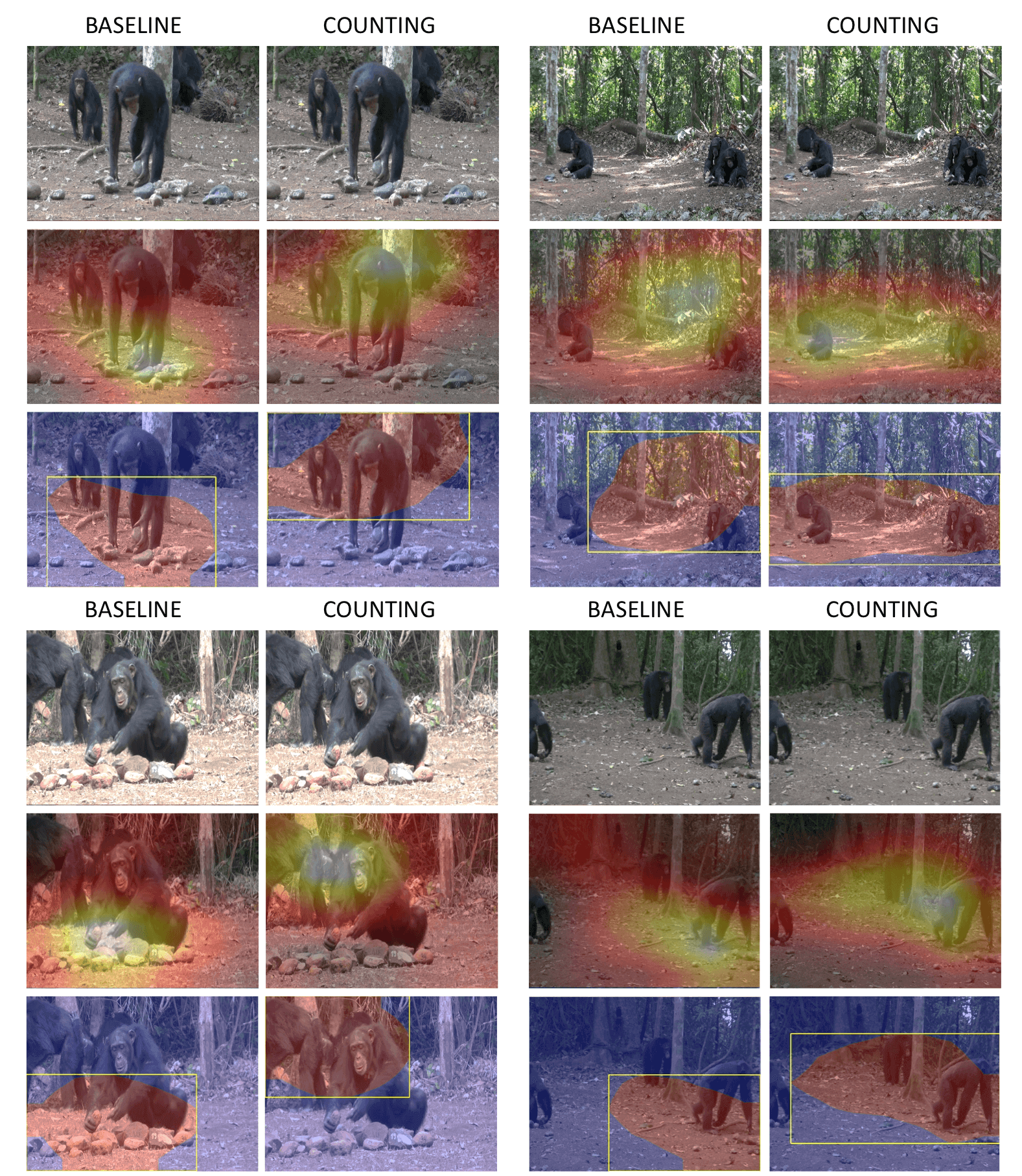}
    \caption{Region proposals for both the baseline (recognition) and counting method. Note how the baseline method mistakenly focuses on the background features, rocks and trees, to recognise individuals.}
    \label{fig:counting_props}
    \vspace{-0.7em}
\end{figure}

\noindent\textbf{Why use counting to localise?}
Our method begs the following question: if a model must identify discriminative regions to be able to count individuals, surely it must also identify these regions to perform fine-grained recognition? In this case we could just train the fine grained recognition network to obtain region proposals, crop regions and then retrain the recognition network in an iterative manner. 
However, counting objects is a much easier task than the fine-grained recognition of identities (a widely studied phenomenon in psychology, called subitizing~\cite{clements1999subitizing} suggests that humans are able to count objects with a single glance if the total number of objects is small). We find that this leads to much better region proposals, as demonstrated in Figure~\ref{fig:counting_props} where we show proposals obtained from a counting model and from an identification model. By tackling an easier task first, our model is using a form of curriculum learning~\cite{bengio2009curriculum}.

\vspace{-0.2em}
\section{Face and Body Tracking and Recognition}
\begin{figure*}
\centering
  \includegraphics[width=1\linewidth]{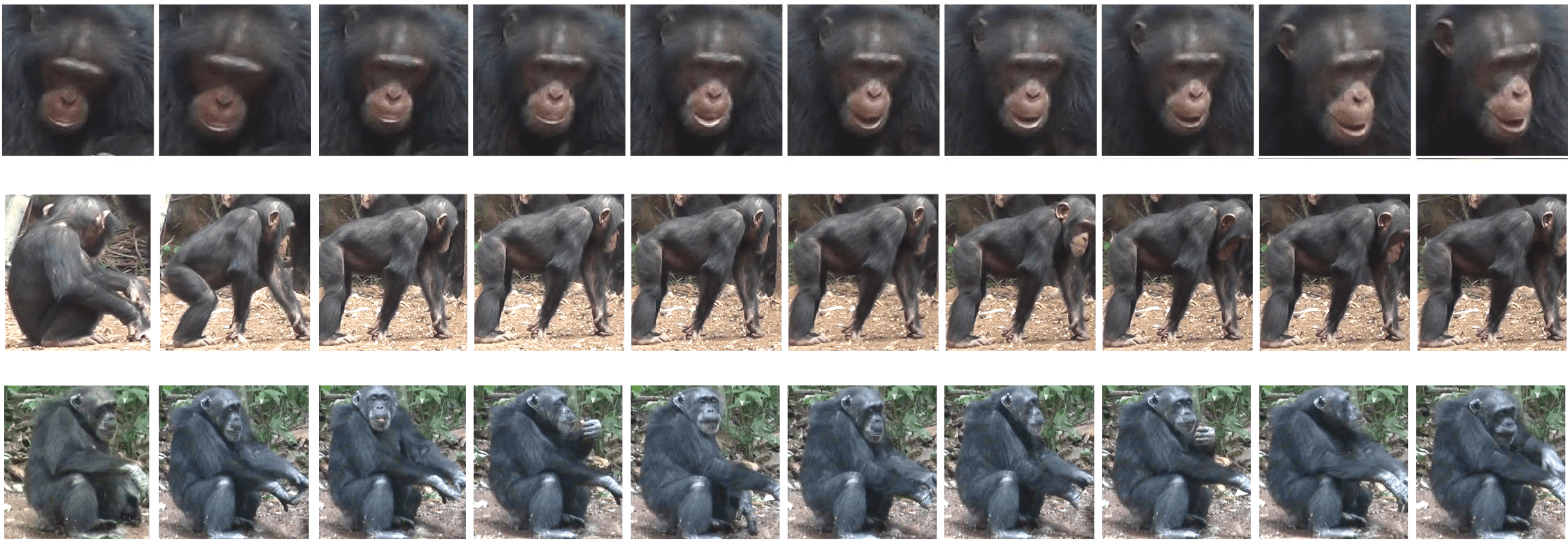}
  \caption{Chimpanzee tracks for face (top row) and body (bottom two rows).}
\label{fig:tracks}
\end{figure*}

\vspace{-0.2em}
\label{sec:tracks}
In order to test recognition methods that explicitly use only face and body regions, we first create a chimpanzee face and body detection dataset, by annotating bounding boxes using the VIA annotation tool~\cite{dutta2016via}. We then train a detector with these detection labels, and run the detector on every frame of the video. A tracker is then run to link up the detections to form \textit{face-tracks} or \textit{body-tracks}, which then become a single unit for both labelling and recognition. Examples are shown in Figure~\ref{fig:tracks}. Finally, we train a standard CNN multi-class classifier on the regions in the track using a cross-entropy loss on the identities in the dataset to train a recognition model.

\FloatBarrier

\section{Datasets}
\begin{figure}
    \centering
    \includegraphics[width=1\columnwidth]{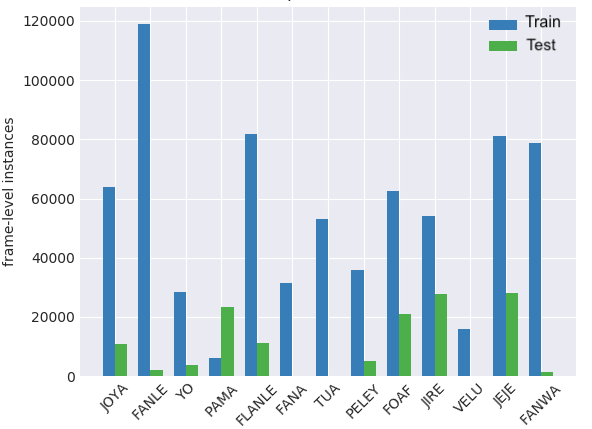}
    \caption{Instance frequency histograms for each individual in the Chimpanzee Bossou dataset.}
    \label{fig:Chmp_class_freq}
\end{figure}

\noindent\textbf{Chimpanzee Bossou Dataset:}
We use a large, un-edited video corpus of chimpanzee footage
collected in the Bossou forest, southeastern Guinea, West
Africa. Bossou is a chimpanzee field site established by
Kyoto University in 1976
~\cite{sugiyama1984population,humle2011location,matsuzawa2011field,FaceBossou}. 
Data
collection at Bossou was done using multiple cameras to document chimpanzee behaviour at a natural forest clearing (7m x 20m) located
in the core of the Bossou chimpanzees' home range. The videos
were recorded at different times of the day, and span a range of
lighting conditions. Often there is heavy occlusion of individuals due to trees and other foliage. The individuals move around and interact freely with one
another and hence faces in video have large variations in scale, motion blur and
occlusion due to other individuals. Often faces appear as extreme
profiles (in some cases only a single ear is visible). 
While the original Bossou dataset is a massive archive with
over 50 hours of data from multiple years, in this paper we use
roughly 10 hours of video footage from the years 2012 and 2013, of
which we reserve 2 hours for testing. Chimpanzees are visible for
the vast majority of this footage, therefore we also include sampled
frames of just the forest background ($n=0$) from other years to permit negative
training for all methods.\\
\\
\noindent\textbf{Dataset annotation and statistics:}
We manually provide frame level annotations (i.e.\  name tags for the individuals present) 
for every frame in the videos using
the VIA video annotation
tool~\cite{dutta2019vgg}. VIA is an open source project based solely on HTML, Javascript and CSS (no dependency on external libraries) and runs entirely in a web browser\footnote{\url{http://www.robots.ox.ac.uk/~vgg/software/via/}}.
In addition, we compute face and body detections and tracks (as described in
Section~\ref{sec:tracks}) and also label these tracks  manually using the VIA tool. All identity labelling was done by an expert anthropologist familiar with the identities in the archive.
The statistics of the dataset
are given in Table~\ref{tab:chimpstats}. The frame-level frequency histogram for each individual is shown in Figure~\ref{fig:Chmp_class_freq}, where an instance of an individual is defined as a frame for which the individual is visible.

\begin{table}
    \centering 
    \small
    \begin{tabular}{l | c c c }
        & hours & \#frames & \# individuals \\
        \hline \hline
        train & 8.30 & 830k & 13 \\
        test & 1.56 & 161k & 10 \\ 

        total & 9.86 & 992k & 13 \\
    \end{tabular}
    \caption{\small{
    Dataset Statistics for the Chimpanzee Bossou dataset.
    We annotate facetracks, bodytracks and identities at a frame level. }}
    \label{tab:chimpstats} 
\end{table}

\section{Experiments}~\label{sec:results} 
We first evaluate the performance of the face-track and body-track methods, in particular the proportion of frames
that they can label (the frame recall), 
and their identity recognition performance. This is then compared to the performance
of the frame-level CCR method using average precision (AP) to analyse the trade-offs
thoroughly. We also compare the CCR method to a simple baseline, where an identity
recognition CNN is trained directly on the resized raw (not zoomed in) images $\rvx'$.\\
\vspace{-0.5em}
\subsection{Evaluation Metrics}
 \vspace{-0.1em}
\noindent\textbf{Detector Recall:} 
The detector recall is the proportion of instances where faces (or bodies) are detected and tracked. This provides
an upper bound on the  number of individual instances that can be
recognised from the video dataset using the face-track or body-track methods.
We note that this is a function of two effects: (1) the visibility of the
face or body in the image (faces could be turned away,
be occluded etc);  and (2) the performance of the detection and tracking
method (i.e.\ is the face detected even if it is visible); though we do not distinguish these two effects here. \\
\\
\noindent\textbf{Identification Accuracy:} 
This is the proportion of detections that are labelled correctly (each face-track or body-track can only be one of
the possible identities).\\


\noindent\textbf{System-level Average Precision (AP):} 
For the face (and body) track methods, the precision and recall for each individual is computed as follows:
all  tracks are ranked by the score of the individual face classifier;
if the track belongs to that individual, then all the frames that contain that track are counted as recalled;
if the track does not belong to that individual, then the frames that contain that track are not recalled 
(but the precision takes these negative tracks into account), 
i.e.\  we only recall the frames containing a track if we correctly identify the individual in that track.
For the frame level CCR method, the frames are ranked by the frame-level identity classifier, and
the precision and recall computed for this ranked list. We then calculate both the micro and macro Average Precision score over all the individuals. Macro Average Precision (mAP) takes the mean of the AP values for every class, whereas Micro Average Precision (miAP) aggregates the contributions of all classes to compute its average metric. For our heavily class unbalanced datasets, the latter is a much better indicator of the overall performance.
(histograms are provided in the supplementary material). 
 
\subsection{Implementation Details} \label{sec:implement}
 
\noindent\textbf{CNN architecture and training:}
For a fair comparison, we use the following hyperparameters across
\textit{all} recognition models: a ResNet-18~\cite{he2016deep}
architecture pretrained on ImageNet \cite{imagenet_cvpr09} with input size $H',W' = 224$ i.e.\ for the counting
CNN $ c_{\theta}(\rvx')$, the fine-grained identity CNN
$R_{\phi}(\rvx_{crop}')$, and the recognition CNNs used for both the
body and the face models. This architecture achieves a good trade-off between
performance and number of parameters. In principle any deep CNN
architecture could be used with our method.  The models are trained
and tested on every third frame from the video
(to avoid the large amount of redundancy in consecutive
frames).
We use a batch size of 64; standard data augmentation (colour jittering, horizontal
flipping etc.) but only random cropping on the negative ($n=0$) samples.
All models are trained end-to-end in PyTorch~\cite{paszke2017automatic}. Models and code will be released. \\
\\
\noindent\textbf{Face and Body tracks:}
The face and body tracks were obtained by
training a Single Shot MultiBox Detector (SSD) \cite{SSD}, on 8k and
16k bounding box annotations respectively. The annotations were gathered on frames
sampled every 10 seconds from a subset of training footage as well as
from videos from other years.
The detectors are trained in PyTorch with 300 $\times$ 300 input
resolution and the same data augmentation techniques as \cite{SSD}. We
use a batch size of 64 and train the detectors for 95k iterations with
a learning rate of $1E-4$ . We used the KLT
\cite{lucas1981iterative,tomasi1991tracking} and SPN
\cite{Li_2018_CVPR} tracker to obtain face and body tracks
respectively. During the recognition stage, predictions are averaged
across a track. \\
\\
\noindent\textbf{Count, Crop and Recognise:}
The  coarse-grained counting CNN is applied on the entire
dataset and the CAM of the highest softmax prediction for each
image recorded. The CAMs, just $7  \times 7$ int arrays, are saved cheaply as
grey-scale images each of size 355 bytes. Alternatively, this can be
performed online during training, albeit at a greater computational
cost since the CAMs are recomputed every epoch.  Before training the
recognition stage, we upsample the CAMs to the size of its
corresponding image and threshold with $T=0.5$,  perform full-resolution cropping and then resize back to $224 \times 224$, the input size of the fine-grained identity CNN $R_{\phi}(x_{crop})$. Fine-grained recognition
is then performed on these cropped regions. 

\subsection{Results}~\label{sec:exp} 
\vspace{-2.7em}
\paragraph{Detector recall and identification accuracy:}
The performance is given in Table~\ref{table:detectorchimp}. It is
clear that recall is a large limitation for both the face-track and body-track methods. The face detector recall is low (less than 40\%), far lower than that of the body detector. This reflects
the fact that the chimpanzee's faces are not visible in many frames, rather than failures of the face detector.  Hence
even a perfect face recognition system would miss many chimpanzee instances at the frame level.
While the identification accuracy for chimpanzees, is slightly higher for faces than for bodies, the relatively high recall of the body-track method shows a clear advantage over faces.

\begin{figure}
 {\centering
\begin{minipage}{0.49\columnwidth}
   \small{
\begin{tabular}{lll}
\multicolumn{1}{l}{Method} & mAP & miAP \\ \hline \hline
Random                                      & 28.4  & 29.2  \\  
Face                                        & 40.1 & 47.1 \\ 
Body                                        & 42.4 & 58.3 \\  \hline \hline
Frame Level                           & & \\
Baseline                                    & 45.5 & 48.2  \\
\textbf{CCR}                     & \textbf{50.0} & \textbf{59.1} \\ 
\hline
\end{tabular}
\label{table:results}
}
\end{minipage}
\vspace{-0.2em}
\begin{minipage}{0.49\columnwidth}
\includegraphics[width=1\linewidth]{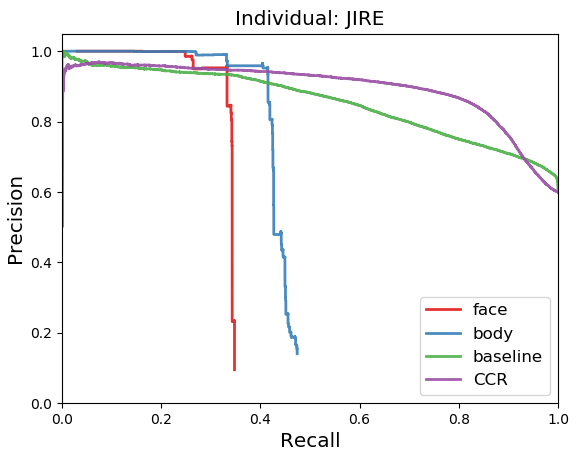}
\end{minipage}

\caption{\small{Left: Comparison of system level AP for all 
methods on the \textit{test} set; Right: PR curves for a single individual from the Chimpanzee dataset.} } \label{fig:results}
}
\vspace{-1.1em}
\end{figure}

\paragraph{System level AP:} Results are given in Figure~\ref{fig:results}, left.
We compare our CCR method to a simple baseline without the Count and Crop stages. 
CCR outperforms the baseline by a large-margin
 (more than 9\% AP). 
The PR curve for the
chimpanzee `JIRE' (Figure~\ref{fig:results}, right),  reiterates the results
that  face-track recall is the lowest, albeit with the
highest precision. In contrast, the CCR method has far higher recall and with a similar level of precision.
The overall AP values (Figure~\ref{fig:results},
left) show that the body-track AP is quite high, since it achieves a
large boost in recall over the face-tracks with
a very small drop in identification accuracy (less than 1\%). We note
that the  CCR method, however,  outperforms the body-track method
as well. This is an impressive performance considering CCR requires
only frame level supervision in training,  and eschews the need to
train a body detector.



   
\begin{table}
    \centering 
    \begin{tabular}{l | c c c c} 
       \multicolumn{1}{c}{} & \multicolumn{4}{c}{} \\
        & \#instances & \#tracks & recall (\%) & test acc. (\%) \\ 
        \hline \hline
        face & $1.02m$ & $5k$ & $39.9$ &71.3 \\ 
        body & $1.64m$  & $12k$ & $64.0$ &70.5 \\ 
        frame  & $2.13m$ & - & $100.0$ & - \\ 
    \end{tabular}
    \caption{Face and body detector recall and identification test accuracy (acc.) results for the Bossou dataset. Recall is calculated as a percentage of the total number of instances annotated at a frame level, which we note as a theoretical upper bound of 100\%. }
    \label{table:detectorchimp} 
\end{table}

\section{Weakly Supervised Localisation of Individuals}

\begin{figure}
    \centering
    \includegraphics[width=0.78\columnwidth]{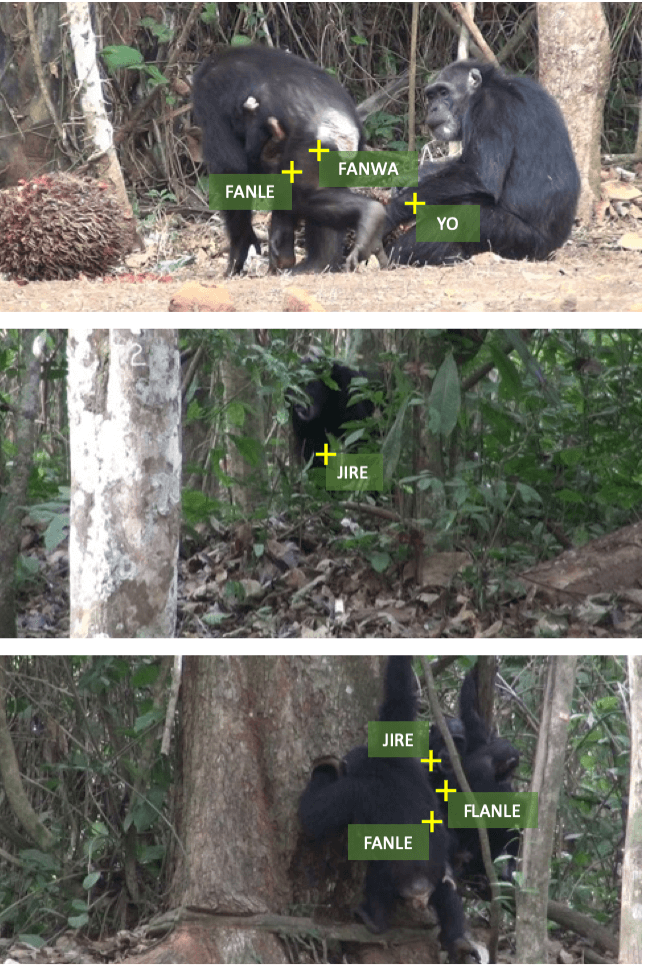}
    \caption{Weakly supervised localisation of individuals.}
    \label{fig:weak_loc}
    \vspace{-1.4em}
\end{figure}

Labelling individuals within a frame offers insight into social relationships by monitoring the frequency of co-occurrences and locations of the capturing cameras. However, unlike face and body detection, the frame level approach does not explicitly localise individuals within the frame, preventing analysis of the local proximity between individuals. To tackle this, we propose an extension to CCR which localises individuals without any extra supervisory data. This is shown in the examples of Figure~\ref{fig:weak_loc}.

Following a similar process to the `Crop' stage in CCR, bounding boxes are generated for each labelled individual from CAMs extracted from the recognition model $R_{\phi}(\rvx'_{crop})$. The locations of the individuals are assumed to be at the centroid of these bounding boxes, with qualitatively impressive results even when the individuals are grouped together.

\section{Interpretability}
In this penultimate section, we introduce a high-granularity visualisation tool to understand and interpret the predictions made by the face and body recognition models.
These tease out the discriminative features learnt by the model
for this task of fine-grained recognition of individuals. Understanding these features can provide new insights to human researchers.


A Class Activation Map (CAM)~\cite{zhou2016learning}, introduced in Section~\ref{sec:ccr}, can be used to localise discrimnative features but it does so at low resolution and thus cannot identify high-frequency features, such as edges and dots. An alternative visualisaton method is Excitation Backprop (EBP)~\cite{excitation_backprop}. EBP achieves high-granularity visualisation via a top-down attention model, working its way down from the last layer of the CNN to the high resolution input layer. Activations are followed from layer to layer with a probabilistic Winner-Take-All process.

\begin{figure}
\begin{subfigure}[b]{1\columnwidth}
  \centering
  \includegraphics[width=0.49\textwidth, height=2cm]{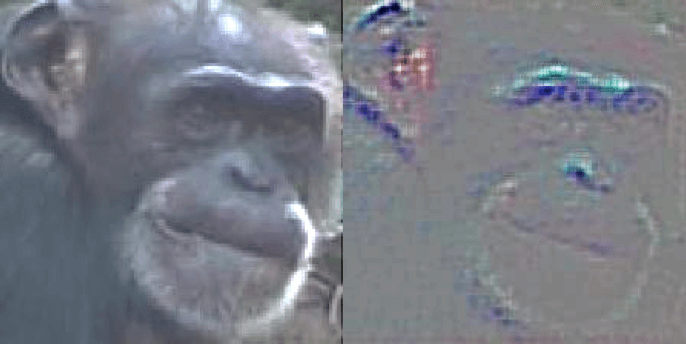}  
  \includegraphics[width=0.49\textwidth, height=2cm]{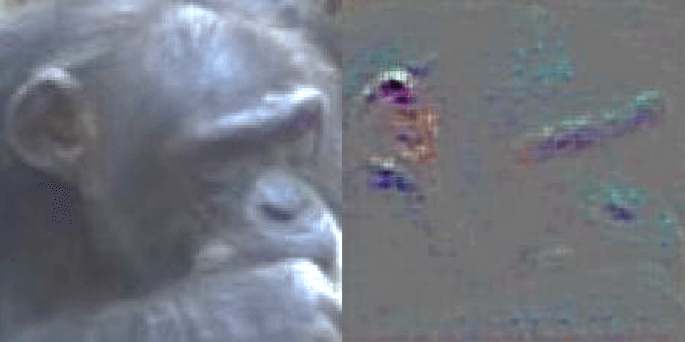}  
  \caption{Fana - Face}
  \label{fig:ebp_fana}
\end{subfigure}
~
\begin{subfigure}[b]{1\columnwidth}
  \centering
  \includegraphics[width=0.49\textwidth, height=2cm]{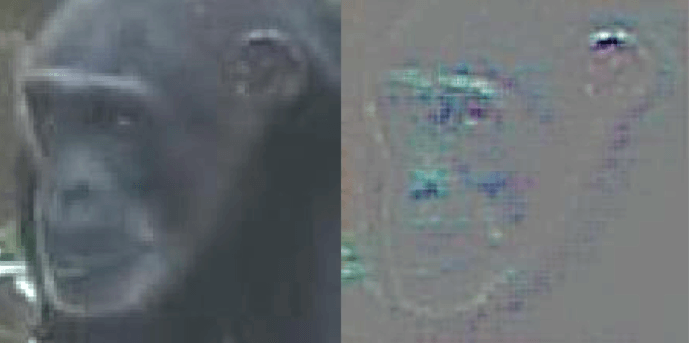}
  \includegraphics[width=0.49\textwidth, height=2cm]{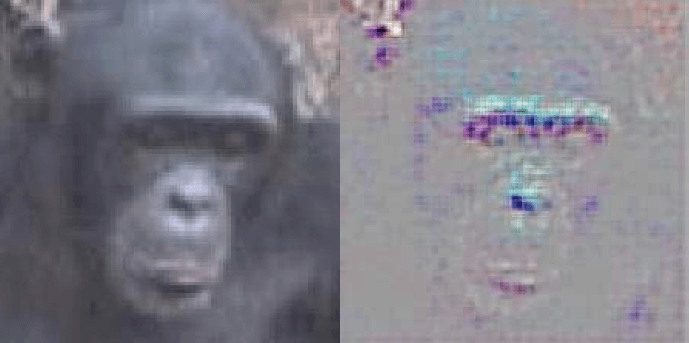}  
  \caption{Pama - Face}
  \label{fig:ebp_pama}
\end{subfigure}
~
\begin{subfigure}[b]{1\columnwidth}
  \centering
  \includegraphics[width=0.49\textwidth, height=2cm]{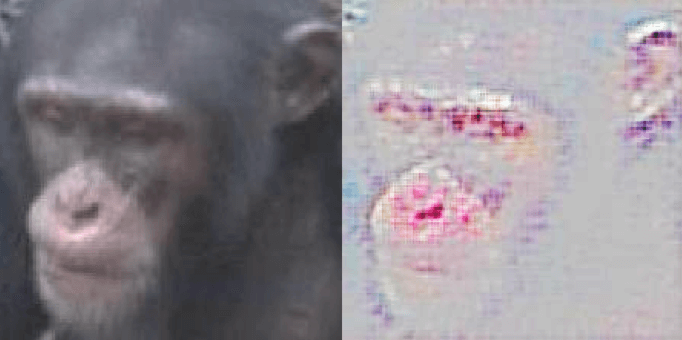}
  \includegraphics[width=0.49\textwidth, height=2cm]{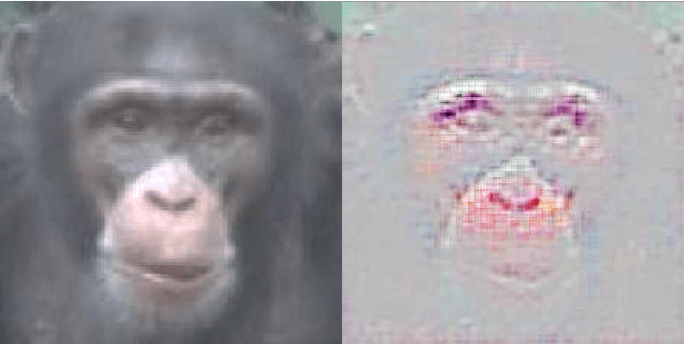}  
  \caption{Jeje - Face}
  \label{fig:ebp_jeje}
\end{subfigure}
~
\begin{subfigure}[b]{1\columnwidth}
  \centering
  \includegraphics[width=0.49\textwidth, height=2cm]{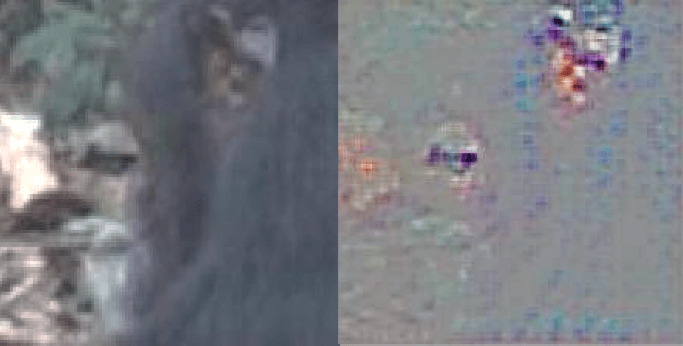}
  \includegraphics[width=0.49\textwidth, height=2cm]{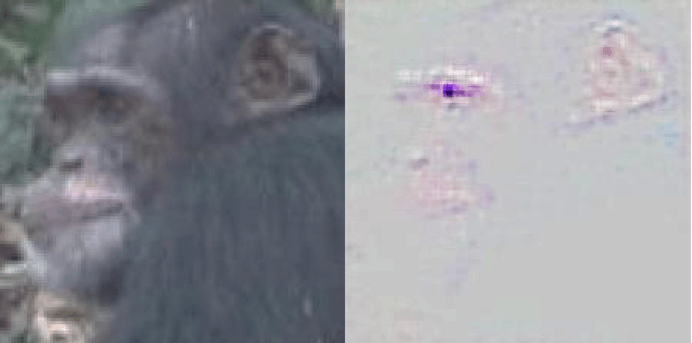}  
  \caption{Foaf - Face}
  \label{fig:ebp_foaf}
\end{subfigure}
~
\begin{subfigure}[b]{1\columnwidth}
  \centering
  \includegraphics[width=0.49\textwidth, height=2cm]{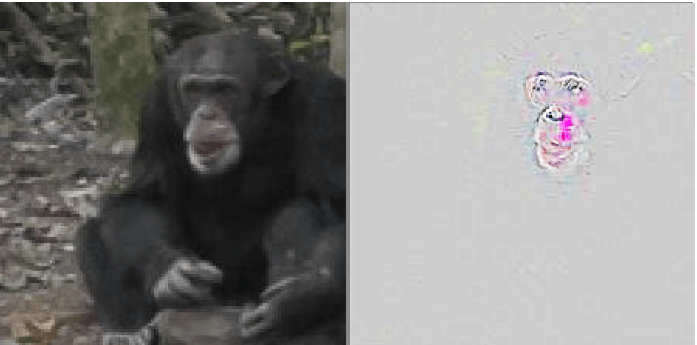}  
  \includegraphics[width=0.49\textwidth, height=2cm]{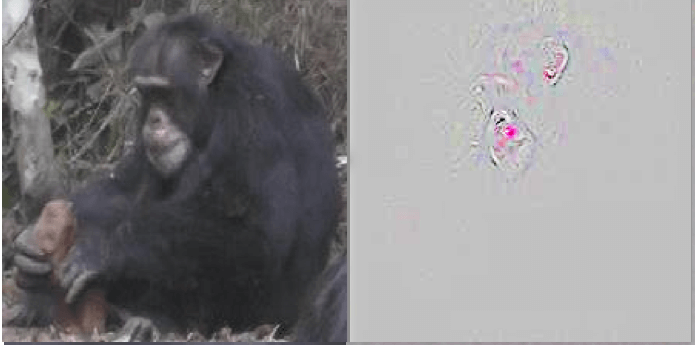}  
  \caption{Foaf - Body}
  \label{fig:ebp_foaf_b}
\end{subfigure}
~
\begin{subfigure}[b]{1\columnwidth}
  \centering
  \includegraphics[width=0.49\textwidth, height=2cm]{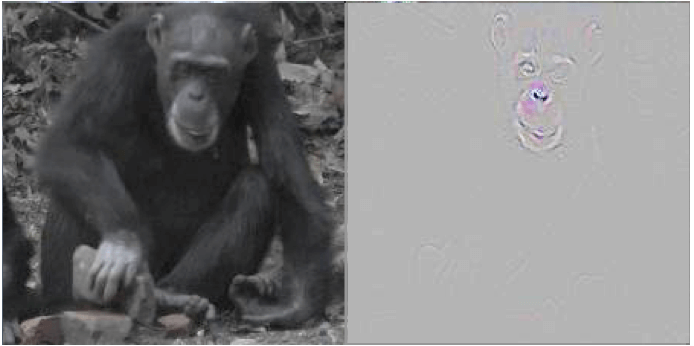}
  \includegraphics[width=0.49\textwidth, height=2cm]{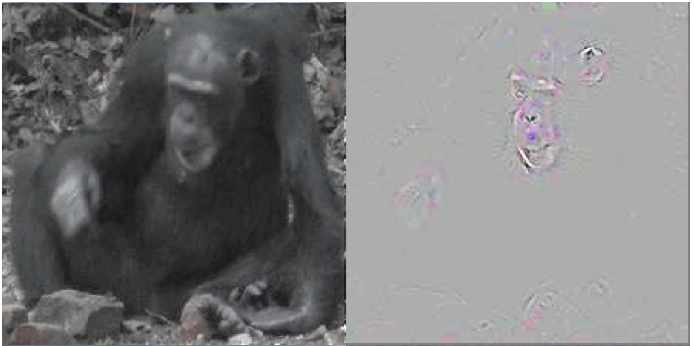}  
  \caption{Fana - Body}
  \label{fig:ebp_fana_b}
\end{subfigure}
~
\begin{subfigure}[b]{1\columnwidth}
  \centering
  \includegraphics[width=0.49\textwidth, height=2cm]{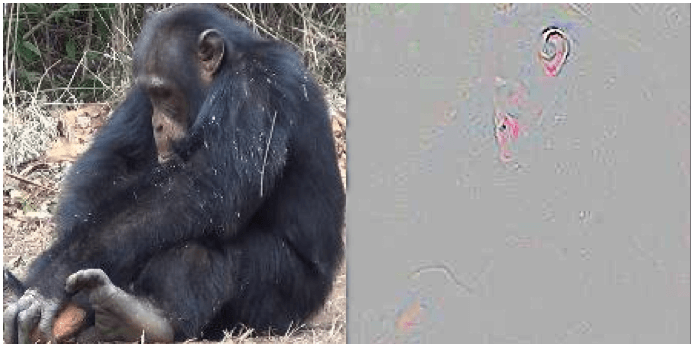}
  \includegraphics[width=0.49\textwidth, height=2cm]{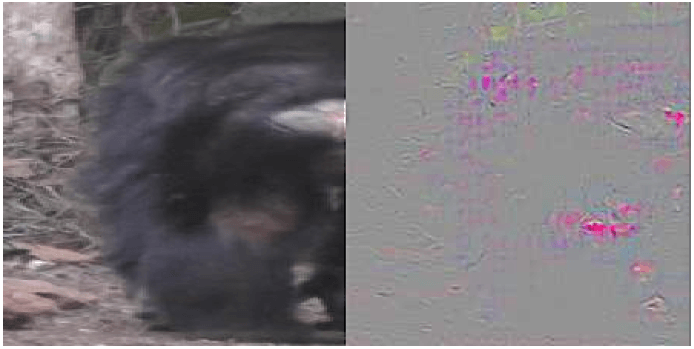}  
  \caption{Jeje - Body}
  \label{fig:ebp_jeje_b}
\end{subfigure}
\caption{Excitation Backprop~\cite{excitation_backprop} visualisations (right) from the face and body recognition model for example images of of individuals in the Chimpanzee Bossou dataset.}
\label{fig:ebp}
\vspace{-1em}
\end{figure}

In Figure~\ref{fig:ebp}, we show the EBP visualisations from the face recognition model of example images of individuals in the Bossou dataset. When the ears are visible, the face model shows high activation on the ear region -- similarly for the brow and mouth regions. Upon closer inspection of the original face images, the ears of each individual are indeed highly unique and distinguishable. The expert anthropologist, who manually labelled the dataset, noted that he doesn't pay particular attention to the ears when identifying the individuals. Perhaps our discovery of ear uniqueness in chimpanzees in this dataset, and possibly all chimpanzees, could improve expert's recognition of chimpanzee individuals.

The EBP visualisation for the body recognition model in Figure~\ref{fig:ebp} reiterates the importance of the face and ears in distinguishing the individuals. Further, note Jeje’s hairless patch on his left leg in the top of Figure~\ref{fig:ebp_jeje_b} and corresponding EBP activation, indicating that the body recognition model also uses distinguishing marks on the body. Similarly, Foaf’s white spot above his upper lip (Figure~\ref{fig:ebp_foaf_b}) is another region of high activation. The presence of the white spot was unbeknownst to the anthropologist who noted he would now use this information to identify Foaf in the future. These two examples show that a CNN’s learned discriminative features for a specific individual can be visualised and interpreted by humans. Of course, these findings are not statistically relevant and quantitative analysis would be needed in order to determine the effectiveness of the use of recognition CNNs to train human experts.


\section{Conclusion}
We have proposed and implemented a simple pipeline for fine-grained recognition of individuals using only frame-level supervision. This has  shown that a counting objective allows us to learn very good region proposals, and zooming into these discriminative regions gives substantial gains in recognition performance. Many datasets `in the wild' have the property that resolution of individuals can vary greatly with scene depth, and with cameras panning and zooming in and out. Our frame-level method approaches the precision of face-track and body-track recognition methods, whilst now allowing a much higher recall. We hope that our newly created dataset will spur further work in high-recall frame-level methods for fine-grained individual recognition in video, and that our preliminary work on interpretability of CNNs for classifying individuals of species gives insight on identifying discriminative features.
\\
\\
\textbf{Acknowledgments:} This project has benefited enormously from discussions with Dora Biro and Susana Carvalho at Oxford. We are grateful to Kyoto University's Primate Research Institute for leading the Bossou Archive Project, and supporting the research presented here, and to IREB and DNRST of the Republic of Guinea.
This work is supported by the EPSRC programme grant Seebibyte EP/M013774/1.  A.N.  is funded by a Google PhD fellowship; D.S. is funded by
the Clarendon Fund, Boise Trust; Fund and Wolfson College, University
of Oxford.

We also thank Dr Ernesto Coto, his assistance was paramount to the success of this work.




\clearpage
\bibliographystyle{ieee}
\bibliography{shortstrings,vgg_local,vgg_other,egbib}

\end{document}


\maketitle
\section{Class Histograms} 
\begin{figure}[h]
\centering
\begin{subfigure}{\includegraphics[width=0.45\linewidth]{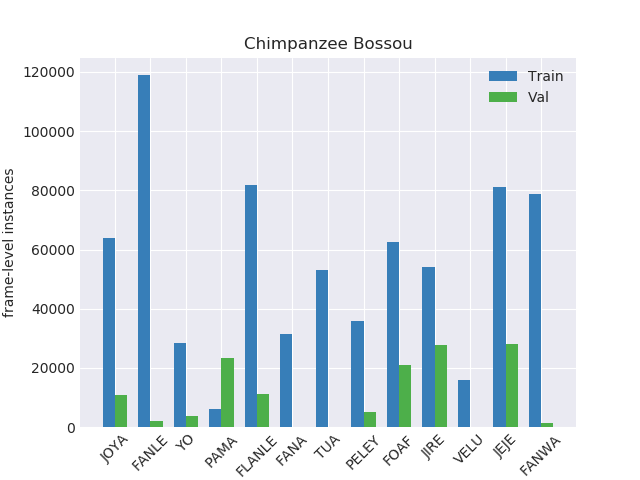}}
\end{subfigure}
\begin{subfigure}{\includegraphics[width=0.45\linewidth]{images/appendix/sherlock_freq.png}}
\end{subfigure}
\caption{Dataset frame-level instances for Chimpanzee and Sherlock TV datasets}
\label{fig:class_freq}
\end{figure}

Both the datatsets used in this work are highly unbalanced. Class frequencies for all individuals can be seen in Fig. \ref{fig:class_freq}.

\vspace{-0.5em}
\section{Micro-Averaged Precision Recall Results}
Micro-Averaged PR curves for the all the individuals in both datasets can be seen in Fig. \ref{fig:PR}. The AP for both these graphs has been given in Fig. 5 of the main paper.
\begin{figure}[h]
\centering
\begin{subfigure}{\includegraphics[width=0.45\linewidth]{images/appendix/chimp_pr.png}}
\end{subfigure}
\begin{subfigure}{\includegraphics[width=0.45\linewidth]{images/appendix/sherlock_pr.png}}
\caption{The micro-averaged Precision Recall for the Chimpanzee (left) and Sherlock TV (right) dataset.}
\end{subfigure}
\label{fig:PR}
\end{figure}
\section{Localisation using Counting vs Identity CNN} 
In this section we compare region proposals obtained by using the proposal method described in Sec. 3 of the main paper for both the counting CNN and the recognition CNN. The proposals are shown in Fig. \ref{fig:proposals}. It is clear that counting is a much better learning objective for this task, as the baseline method mistakenly focuses on the background features, rocks and trees.
 \begin{figure}[h]
     \centering
     \includegraphics[width=0.95\linewidth]{images/appendix/CAM_count_vs_baseline.png}
     \caption{Region proposals for both the baseline and counting method. Note how the baseline method mistakenly focuses on the background features, rocks and trees, to recognise individuals.}
     \label{fig:proposals}
 \end{figure}
 \clearpage
\section{Screenshots of Annotation Tools} 
A screenshot of the VIA tool \cite{dutta2016via} for frame level annotation of the chimpanzee dataset can be seen in Fig. \ref{fig:screenshot}.
 \begin{figure}[h]
     \centering
     \includegraphics[width=0.95\linewidth]{images/appendix/VIA_video.png}
     \caption{Screenshot of the VIA tool \cite{dutta2016via} for frame level annotation of the chimpanzee dataset.}
     \label{fig:screenshot}
 \end{figure}
 \clearpage
 \bibliography{shortstrings,vgg_local,vgg_other,egbib}